\pdfoutput=1

\documentclass[11pt]{article}

\usepackage[final]{acl}

\usepackage{times}
\usepackage{latexsym}

\usepackage[T1]{fontenc}

\usepackage[utf8]{inputenc}

\usepackage{microtype}

\usepackage{inconsolata}

\usepackage{graphicx}
\usepackage{booktabs} 
\usepackage{caption}
\usepackage{algorithm}
\usepackage{algorithmic}
\usepackage{balance}

\title{Interpretable Cross-Examination Technique (ICE-T): Using highly informative features to boost LLM performance}

\author{Goran Muric \\
  InferLink Corporation \\
  Los Angeles, California \\
  \texttt{gmuric@inferlink.com} \\\And
  Ben Delay \\
  InferLink Corporation \\
  Los Angeles, California \\
  \texttt{bdelay@inferlink.com} \\\And
  Steven Minton \\
  InferLink Corporation \\
  Los Angeles, California \\
  \texttt{sminton@inferlink.com} \\}

\begin{document}
\maketitle
\begin{abstract}
In this paper, we introduce the Interpretable Cross-Examination Technique (ICE-T), a novel approach that leverages structured multi-prompt techniques with Large Language Models (LLMs) to improve classification performance over zero-shot and few-shot methods. In domains where interpretability is crucial, such as medicine and law, standard models often fall short due to their ``black-box'' nature. ICE-T addresses these limitations by using a series of generated prompts that allow an LLM to approach the problem from multiple directions. The responses from the LLM are then converted into numerical feature vectors and processed by a traditional classifier. This method not only maintains high interpretability but also allows for smaller, less capable models to achieve or exceed the performance of larger, more advanced models under zero-shot conditions. We demonstrate the effectiveness of ICE-T across a diverse set of data sources, including medical records and legal documents, consistently surpassing the zero-shot baseline in terms of classification metrics such as F1 scores. Our results indicate that ICE-T can be used for improving both the performance and transparency of AI applications in complex decision-making environments.
\end{abstract}

\section{Introduction}There are numerous prompting strategies to achieve good performance using generative Large Language Models (LLMs). Take, for instance, a binary classification problem, where a system should classify the given text into one of two classes. A typical zero-shot approach is to prompt the model with a given text and carefully designed question, that will yield an appropriate answer. There are also multiple variations on that approach that include ``chain-of-thought'' prompting~ \cite{wei2022chain,wang2022self,kojima2022large}, ``few-shot learning''~ \cite{schick2022true,gu2021ppt}, ``self-instruct''~\cite{wang2022self_instruct,yang2024gpt4tools} prompting and ``iterative refinement''~\cite{wu2022ai, trautmann2023large}. These tactics are used to get a better sense of the model's underlying reasoning or to surpass the performance achieved by the standard zero-shot method.

These options are usually used in cases where using highly specialized fine-tuned LLMs is not a viable option because it is often of utmost importance to understand how decisions are made. This is especially true in fields like medicine, where decisions based on opaque, ``black-box'' models are usually not acceptable. Although zero-shot or few-shot prompting methods can potentially offer explanations for their reasoning, these explanations are often unstructured and lack quantifiability. On the other hand, while finely tuned models may achieve superior performance, they frequently struggle to articulate the rationale behind their outputs unless explicitly trained for this purpose, a process that is labor-intensive. Additionally, outputs from such models may also suffer from the lack of structured reasoning representation.

In cases where using ``black-box'' models is not practical, and where interpretability is important, users have the option to develop a structured reasoning process by asking several questions to achieve a desired output. There are three main problems that arise with this approach: 1) Non-experts have little chance to develop a good set of questions and rules that ensure optimal model performance; 2) Designing an accurate rule set becomes challenging since individual instances may not perfectly align with all desired criteria, resulting in a mix of positive and negative responses to different rules; and 3) The potential combinations of these rules can become overwhelmingly numerous, making it impractical to hard-code every possible scenario. 

In the paper, we propose a method that attempts to overcome the three issues outlined above. We refer to the method as the Interpretable Cross-Examination Technique, or ICE-T for brevity. Our approach exhibits strong performance,  consistently surpasses the benchmark set by a zero-shot baseline, and also offers a high level of interpretability. The core concept here is that rather than using a single prompt to get a response from an LLM and making a decision based on that single output, we engage the LLM with multiple prompts, covering various questions. We then combine the responses from all these prompts and use the outputs to make a decision. Compared to other methods that are based on multi-prompting, our approach is fundamentally different in the way the decisions are made. Specifically, we take the responses from the LLM, convert them into numerical values to create a feature vector, and then input this vector into a traditional classifier to determine the final outcome. Since, in this process we create a low-dimensional feature vector with highly informative features, we can then use relatively small classifiers to make a decision. 

We established an experimental setup where we tested our Interpretable Cross-Examination Technique on a simple binary classification task. We tested our approach on a set of multiple datasets split on 17 different tasks and we show that:
\begin{enumerate}
    \item ICE-T consistently outperforms the zero-shot baseline model in most classification metrics
    \item Using a smaller model with ICE-T we can achieve comparable or better results than using larger and essentially more capable model with zero-shot approach
\end{enumerate}

Furthermore, this approach can be highly interpretable, allowing experts to clearly understand the rationale behind the decision-making process\footnote{Degree of interpretability may vary depending on the machine learning method selected for the final classification task. The decision on which method to employ should be guided by a consideration of the trade-offs between interpretability and performance tailored to the unique demands of each task}. Additionally, tools commonly used for tabular machine learning can be employed to enhance the understanding of the data. While this technique is specifically evaluated for binary classification within this paper, its applicability potentially extends across a broad spectrum of scenarios.

\subsection{Motivation}
The ICE-T method was initially conceived at InferLink in a commercial consulting project, where we needed to address a complex challenge in biomedical text classification. The project's goals were to develop a model that could perform at a level comparable to human experts, provide interpretable results, and allow for detection of potentially mislabelled data. Initially, conventional ``black-box'' models such like fine-tuned BERT-based ones underperformed, as well as zero-shot or few-shot learning methods using LLMs. This led to the creation of the ICE-T, which improved the performance of classification, while gaining interpretability and allowing for the correction of labeling errors. ICE-T was used initially for the purpose of classifying biomedical data for a specific commercial purpose. While the specifics of this initial task and data remain confidential, we have conducted further testing on additional publicly available datasets and decided to make the method publicly accessible.

\section{Related Work}
Our proposed solution addresses three core aspects of using large language models for inference: prompting, in-context learning, and interpretability. It is built on top of the ever-growing body of knowledge that comes from those domains. 

\subsection{Prompting techniques}
Numerous techniques have been developed to improve the fundamental zero-shot approach. Among these, the ``chain-of-thought'' (CoT) prompting is particularly notable. This method is used to prompt the model to systematically articulate its reasoning process in a step-by-step manner before reaching a conclusion. Research has shown that chain-of-thought prompting improves performance on a range of arithmetic, commonsense, and symbolic reasoning tasks~\cite{cot_NEURIPS2022_9d560961, wei2022chain, wang2022self}. Even simple tweaks such as  adding ``Let's think step by step'' before each answer can significantly outperform zero-shot LLM performances on diverse benchmark reasoning tasks~\cite{kojima2022large,nye2021show}. Such generated chains that prompt language models to break down their reasoning into steps often cause errors in inference time. To reduce these errors, some researchers employ a method known as automatic Chain of Thought  prompting. This technique, which generates demonstrable examples, has proven to be more effective than earlier, simpler CoT approaches~\cite{zhang2022automatic}. Lastly, ``iterative refinement'' involves repeatedly prompting the model with slightly altered versions of the original text or question, honing in on a more accurate or nuanced answer through successive iterations. Each of these strategies can be tailored to the specific needs of a task, leveraging the model's capabilities in different ways to achieve optimal performance. 

Several approaches involve using multiple prompts in a chain, where the output of one step becomes the input for the next, thus aggregating the gains per step~\cite{wu2022ai}, or decomposing complex tasks into smaller, manageable components~\cite{trautmann2023large}. Additionally, ``self-instruct''~\cite{wang2022self_instruct,yang2024gpt4tools} prompting can be used, where the model generates its own instructions or clarifications based on the initial prompt, attempting to refine or better understand the task before generating a response. Another set of approaches uses multiple models or multiple instances of the same model to improve the performance. The additionally trained models, called ``verifiers'' are used to judge the correctness of model completions. At the inference time, the verifiers would select the most likely answer~\cite{cobbe2021training}.

\subsection{In-context learning}
Large Language Models possess the remarkable ability for in-context learning (ICL), in which they acquire knowledge from a few contextual examples either during inference or during training. Numerous studies have shown that through ICL, LLMs can effectively handle a diverse set of complex tasks~\cite{wei2022emergent}. ICL offers several advantages, notably its ease in including human knowledge into LLMs by using various demonstrations and templates~\cite{liu2021makes,wu2022self}. Furthermore, unlike traditional supervised training methods, ICL operates without the need for additional training, significantly lowering the computational costs when using models to solve new tasks~\cite{dong2022survey}.

One of the most recognizable techniques for in-context learning is ``few-shot learning''~\cite{schick2022true,schick2020s,gu2021ppt,perez2021true} during inference\footnote{A different approach to achieving few-shot learning can occur also during the training phase or during fine-tuning.}. Using this approach, the model is provided with a few examples of text and their corresponding labels or desired outputs within the prompt itself. This method teaches the model the context of the decision-making process, improving its accuracy on similar tasks.

Multiple other studies contributed to refining the ICL methods, focusing on automation, ordering, and selection of prompts. Zhou et al. (2022) introduced the Automatic Prompt Engineer (APE), which automates the generation of instructional prompts, significantly reducing manual effort and improving scalability~\cite{zhou2022large}. Simultaneously, Lu et al. (2021) came up with the method to optimize the ordering of prompts. They employed entropy statistics to evaluate and identify the most effective prompt sequences~\cite{lu2021fantastically}. Rubin et al. (2021) and Liu et al. (2021) both contribute to this area but from different perspectives. Rubin et al. (2021) developed a method for efficiently retrieving prompts using annotated data, streamlining the selection process~\cite{rubin2021learning}. On the other hand, Liu et al. (2021) explored strategic selection methods that go beyond random sampling to leverage the few-shot capabilities of LLMs, aiming to enhance the model's performance through example selection~\cite{liu2021makes}. Adding to the discussion on selection strategies, Zhang et al. (2022) approached example selection as a sequential decision problem. They proposed using a reinforcement learning algorithm to discover policies that improve the generalizability of language models~\cite{zhang2022active}. This perspective introduces a dynamic element to the selection process, aligning with the strategies discussed by Rubin and Liu but through an adaptive, policy-driven approach.

\subsection{Model interpretability}
The challenge of interpreting complex decision processes made by LLMs has hindered their application in critical areas like medicine, where there are significant concerns about regulation~\cite{goodman2017european} and safety~\cite{amodei2016concrete}. Furthermore, this difficulty in understanding the workings of large language models (LLMs) and similar neural network models has restricted their use in domains like science and data analysis~\cite{kasneci2023chatgpt}. In such fields, the primary objective is often to derive a reliable interpretation rather than merely to implement an LLM~\cite{singh2024rethinking}.

The expression of uncertainty in language models is crucial for reliable LLM utilization, yet it remains a challenging area due to inherent overconfidence in model responses. Xiong et al. (2023) and Zhou et al. (2024) both highlight the overconfidence issue in LLMs. Xiong et al. question whether LLMs can express their uncertainty, observing a tendency in LLMs to mimic human patterns of expressing confidence~\cite{xiong2023can}. Simlarly, Zhou et al. note that while LLMs can be prompted to express confidence levels, they remain generally overconfident and unable to convey uncertainties effectively, also when providing incorrect responses~\cite{zhou2024relying}. Ye et al. (2022) add that even when LLMs generate explanations, these may not accurately reflect the model's predictions nor be factually grounded in the input, particularly in tasks requiring extractive explanations~\cite{ye2022unreliability}. However, all the research mentioned above note that these flawed explanations can still serve a purpose, offering a means to verify LLM predictions post-hoc.

It is worth mentioning  feature attribution methods, used beyond the LLM realm in multiple deep-learning applications. Feature attributions in machine learning provide a relevance score to each input feature, reflecting its impact on the model’s output. This methodology helps in understanding how and why certain decisions or predictions are made by a model. 

The approaches developed by Lundberg et al. (2017) and Sundararajan et al. (2017) both delve into this topic but offer distinct methodologies and theoretical foundations. Lundberg et al.~\cite{lundberg2017unified} introduced SHAP (SHapley Additive exPlanations), which provides a unified framework for interpreting predictions. SHAP assigns an importance value to each feature for a specific prediction, leveraging the concept of Shapley values from cooperative game theory. In contrast, Sundararajan et al.~\cite{pmlr-v70-sundararajan17a} developed Integrated Gradients, another method focusing on the attribution of predictions to input features of deep networks. Unlike SHAP, which uses Shapley values, Integrated Gradients relies on the integration of gradients along the path from a chosen baseline to the actual input. Complementing these approaches, Ribeiro et al. (2016) proposed LIME (Local Interpretable Model-agnostic Explanations), which aims to make the predictions of any classifier understandable and reliable by learning an interpretable model localized around the prediction~\cite{ribeiro2016should}. 

Another popular method for understanding neural-network representations is probing. Conneau et al. (2018) initially introduced multiple probing tasks designed to capture simple linguistic features of sentences, setting a foundation for understanding how neural networks encode linguistic properties \cite{conneau2018you}. 

Clark et al. (2019) focused primarily on the behavior of attention heads within transformers. They observed that these heads often broadly attend across entire sentences, and that attention patterns in the same layer tend to exhibit similar behaviors. Crucially, their research links specific attention heads to traditional linguistic concepts like syntax and coreference, suggesting a direct relationship between the model's attention mechanisms and linguistic structures~\cite{clark2019does}, although there is an ongoing debate on the explanatory power of attention in neural network~\cite{bibal-etal-2022-attention}. Unlike Clark et al., who examine what the model attends to, Morris et al.~\cite{morris2023text} explore how information is preserved and can be retrieved from embeddings, offering insights into the reversibility and fidelity of the encoding process. Their method involves a multi-step process that iteratively corrects and re-embeds text, demonstrating the ability to recover most of the original text inputs exactly. Belrose et al. (2023) introduced a technique called causal basis extraction, which aims to identify influential features within neural networks~\cite{belrose2023eliciting}. This method stands out by focusing on the causality within network decisions.

In summary, while chain-of-thought prompting can generate errors during inference, requiring complex corrective approaches, in-context learning techniques also face challenges in prompt optimization and efficient retrieval. Furthermore, interpreting large language models remains problematic, exacerbated by models' tendency to exhibit overconfidence and provide unreliable or unverifiable explanations.

\section{Method}
Training the ICE-T system consists of the following steps: 
\begin{enumerate}
    \item \textit{Generating questions}: the process begins by generating a series of questions designed to prompt the Large Language Model (LLM);
    \item \textit{Prompting the LLM}: Previously generated questions are used to prompt the LLM and collect the yes/no answers;
    \item \textit{Verbalizing the answers}: for each instance within the training dataset, responses to prompts are collected and converted into numerical form, thus creating a low-dimensional feature vector for each instance;
    \item \textit{Training a classifier}: Previously obtained vectors, together with their respective labels, are then used to train a classifier
\end{enumerate}

The Inference stage mirrors the training process: the LLM is presented with the same collection of questions. The responses obtained are numerically encoded in the same manner before being processed by the classifier that was trained during the Training stage. Training and inference process is illustrated in Figure~\ref{fig:method_diagram}. Each step is explained below.

\subsection{Generating questions}\label{sec:generating_questions}
To train and use the system, we need to create multiple questions that more closely reflect the core principles behind the initial yes/no question. Those questions should be crafted in a way to uncover some additional details about the problem. 

Consider a use case where an expert is building a classifier to determine eligibility for medical trials based on patient data. In such a scenario, the classifier needs to assess various clinical inclusion criteria, which are typically derived from patient medical records. One of these criteria could be the patient's language proficiency, for instance, whether they speak English. A naive formulation of this question may be to present the question to the LLM in a prompt like the following:

\begin{small}
\begin{verbatim}
  Does this patient speak English according to 
  their medical records. 
  MEDICAL RECORDS: __RECORDS__
\end{verbatim}
\end{small}

where the \verb|__RECORDS__| represents the appended textual medical records.  Determining the answer to that question, which we call the ``primary'' question, may not be easy given the medical records under consideration, requiring an understanding of somewhat subtle indicators that show if a patient actually speaks English. It is highly unlikely that the medical records will directly state the answer to that question. 

However, a series of ``secondary'' questions such as: 

\begin{small}
\begin{verbatim}
  Is there any documentation of the patient
    requiring an interpreter for English
    during medical visits?
  Do the medical records contain notes written
    in English that indicate communication 
    with the patient?
  Are there any written consents or forms
    completed in English by the patient?
  Are there any notations from providers about
    the patient's ability to understand and 
    speak English?
\end{verbatim}
\end{small}

may allow the model to answer directly based on the information already contained in the documents presented to it, while also serving as strong indicators for the primary question. Secondary questions are also yes/no questions.

\begin{figure}
    \centering
    \includegraphics[width=0.95\linewidth]{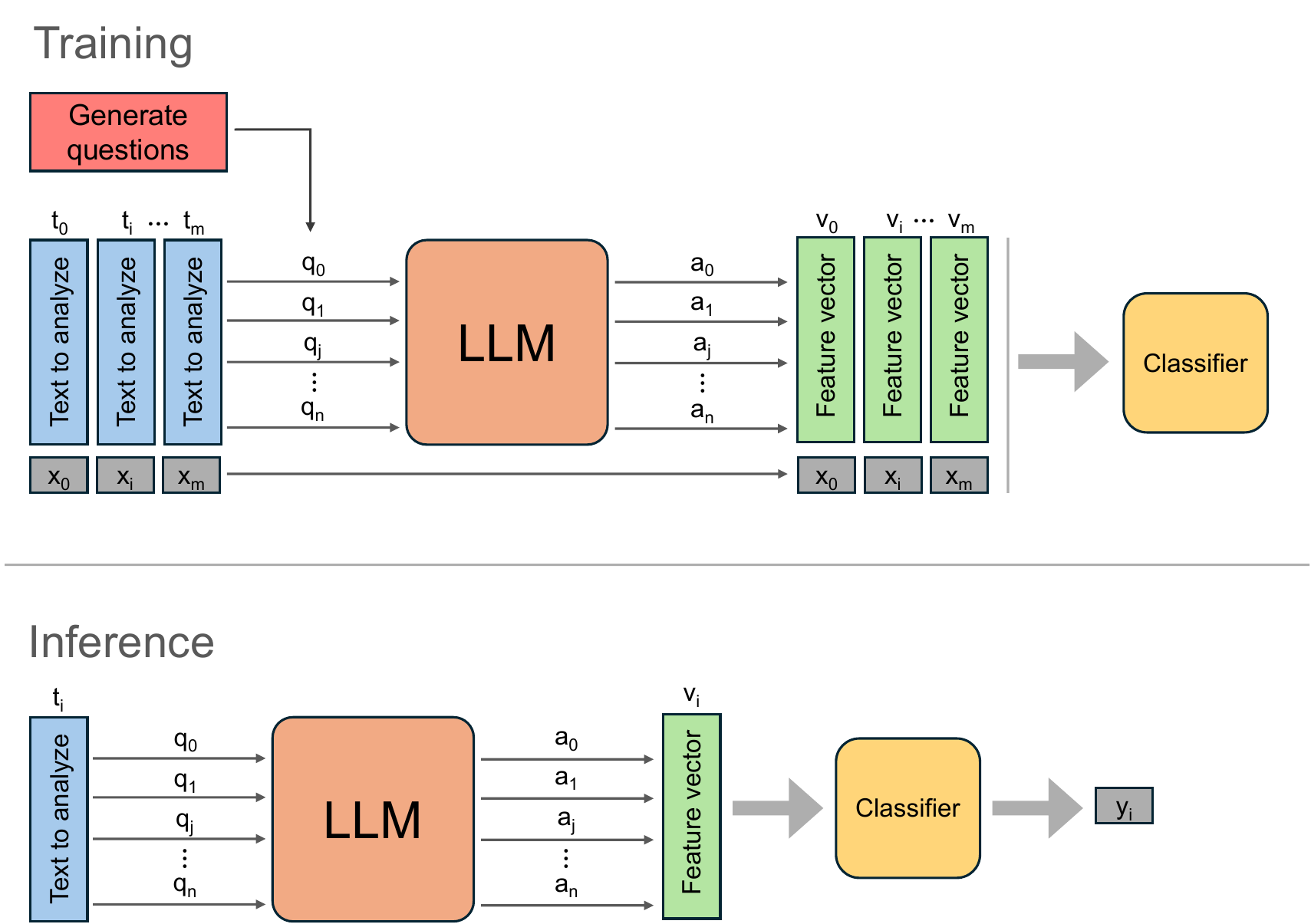}
    \caption{\textbf{Illustration of training and inference process in ICE-T.} In the training phase, the process begins by generating questions to prompt an LLM, which then provides yes/no answers. These answers are verbalized and converted into numerical feature vectors. A classifier is trained using these vectors along with their respective labels. During inference, the LLM is prompted with the same questions, and the answers are similarly processed to predict outcomes using the trained classifier.}
    \label{fig:method_diagram}
\end{figure}

Creating the secondary questions can be done in multiple ways, such as writing the questions manually using the expert knowledge or using the LLM to automatically generate a fixed size set of questions that might be useful in answering the original question. Starting from the primary question $q_{0}$ we generate $n$ additional questions, creating a set of all questions $Q = \{q_0, q_1 \ldots q_n \}$, where $|Q|=n+1$. This process is shown in Figure~\ref{fig:method_diagram} with a red box, illustrating the creation of the questions and using them during the training and inference process. The same set $Q$ of questions is used for both training and inference. 

The number $n$ of secondary questions is decided based on factors such as: number of training samples, availability of the expert knowledge and the level of interpretability needed for a specific task. Our prior small-scale experiments have shown that secondary questions crafted by experts generally lead to improved performance compared to those generated by LLMs. However, in the experiments reported here, we chose a straightforward and reproducible approach where we exclusively use secondary questions created by an LLM. This choice was made to minimize human bias and showcase the method's effectiveness in scenarios where expert input is unavailable. The exact prompts used for creating secondary questions in our experiments are described in Section~\ref{sec:experiments}.

\subsection{Prompting LLM}\label{sec:prompting_llm}
The LLMs are prompted in two occasions. First, they are prompted to obtain the set of secondary questions $Q$, as described in Section~\ref{sec:generating_questions}. Second, for each document, we prompt the LLM with the document and corresponding secondary questions. Then, for each question $q_i$ the output $a_i$ of the LLM is collected, creating a set of outputs for each document. The textual outputs are then assigned a numerical value and transformed into a feature vector $v_i$, through the verbalization process explained in Section~\ref{sec:verbalizing}.

\subsection{Verbalizing the answers}\label{sec:verbalizing}
The output of the LLM in response to each prompt is limited to one of three possible values: Yes, No, or Unknown, depending on the answer to the question posed in the prompt. These responses are subsequently assigned numerical values for analysis, with ``Yes'' translating to 1, ``No'' to 0, and ``Unknown'' to 0.5.

\subsection{Training a classifier}
To train a classifier, we use a set $V$ of low-dimensional numerical vectors, where $|V| = n+1$ and corresponding labels $X$, where each vector $v_i$ has a corresponding binary label $x_i$. Vectors $V$ are obtained from the training textual data after prompting LLM to generate $n+1$ outputs that are then assigned a numerical value. A classifier is then trained using a 5-fold cross-validation process and grid search for the best parameters. A choice of a specific classification algorithm will depend on the size of training data, values distribution and desired performance on a specific classification metric.

\section{Data}\label{sec:data}
This work utilizes data compiled from a range of sources, attempting to include a variety of domains and document lengths. The data used in the experiments described here spans the fields of medicine, law, climate science, and politics. It also includes documents of varying sizes, from brief tweets to extensive legal documents and detailed medical records.

\subsection{Clinical trials}
This dataset comes from Track 1 of the 2018 National NLP Clinical Challenges (n2c2) shared tasks\footnote{https://n2c2.dbmi.hms.harvard.edu/}. It is designed to help in identifying patients within a corpus of longitudinal medical records who either meet or do not meet predefined selection criteria. These criteria are used for determining a patient's eligibility for inclusion in clinical trials.~\cite{Stubbs2019}. The data consists of annotated American English clinical narratives for 288 patients according to whether they met a set of specific criteria. There are 13 criteria in total, and they include: DRUG-ABUSE: Drug abuse, current or past; ALCOHOL-ABUSE: Current alcohol use over weekly recommended limits; ENGLISH: Patient must speak English; MAKES-DECISIONS: Patient must make their own medical decisions; ABDOMINAL: History of intra-abdominal surgery, small or large intestine resection, or small bowel obstruction; MAJOR-DIABETES: Major diabetes-related complication; ADVANCED-CAD: Advanced cardiovascular disease (CAD); MI-6MOS: MI in the past 6 months; KETO-1YR: Diagnosis of ketoacidosis in the past year; DIETSUPP-2MOS: Taken a dietary supplement (excluding vitamin D) in the past 2 months; ASP-FOR-MI: Use of aspirin to prevent MI; HBA1C: Any hemoglobin A1c (HbA1c) value between 6.5\% and 9.5\%; and CREATININE: Serum creatinine $>$ upper limit of normal. For every medical record, each criterion can have one of two potential values: ``met'' or ``not met.'' The value based on whether an individual has fulfilled a particular criterion. Data is split 70/30 on training and test sets respectively. Training test contains 202 medical record while the test set contains 86 records. Note that for some criteria, the ratio between positive and negative class is highly imbalanced. In our analysis we excluded KETO-1YR criterion as it contains no positive samples in the test set and only one positive sample in the training set.\footnote{Our methodology employs classifiers that are trained based on data distributions. As a result, we consistently achieve peak classification metrics, which is not a realistic performance, as the minority class is absent from the test dataset.} 

\subsection{Catalonia Independence Corpus}
This dataset contains a corpus in Spanish that consist of annotated Twitter messages for automatic stance detection~\cite{zotova-etal-2020-multilingual}. It encompasses data collected over a 12-day span in February and March 2019, from tweets originating in Barcelona. Originally, each tweet is categorized into one of three classes: AGAINST, FAVOR, and NEUTRAL. These classes represent the user's stance towards the topic of Catalonia's independence. For the purpose of binary classification and to facilitate more effective comparisons with other datasets, we have omitted the NEUTRAL class, focusing exclusively on the AGAINST and FAVOR categories.

\subsection{Climate Detection Corpus}
This dataset contains climate-related paragraphs extracted from financial disclosures by companies. The text has been collected from corporate annual reports and sustainability reports. The paragraphs from those reports are hand-selected and then annotated as yes (climate-related) or no (not climate-related)~\cite{webersinke2021climatebert}.

\subsection{Medical health advice data}
This dataset comprises a collection of sentences related to the medical domain, each accompanied by a label indicating whether the sentence offers medical advice. The labels can be one of three values: ``strong advice'', ``weak advice'', or ``no advice''.~\cite{yu-etal-2019-detecting} For the purpose of binary classification task we combined ``strong advice'' and ``weak advice'' into a single class: ``advice''. The dataset includes approximately 8,000 samples, which have been divided into training and test datasets following the 80/20 rule.

\subsection{The European Court of Human Rights (ECtHR) Data}
The European Court of Human Rights (ECtHR) hears allegations that a state has breached human rights provisions of the European Convention of Human Rights (ECHR)~\cite{chalkidis-etal-2019-neural}. The dataset for each case includes a series of facts in form of paragraphs extracted from the case description. Additionally, each case is associated with specific articles of the European Convention on Human Rights (ECHR) that may have been violated. In many cases, multiple articles are violated at the same time. To make this a binary categorization problem, we adopted a binary labeling system. Cases are marked with a ``1'' if any ECHR articles are violated, and a ``0'' if no violations are detected.

\subsection{UNFAIR-ToS Dataset}
The UNFAIR-ToS dataset contains 50 relevant on-line consumer contracts, i.e. Terms of Service (ToS) from on-line platforms (e.g., YouTube, Ebay, Facebook, etc.). Each agreement has been annotated at the sentence level to identify various types of potentially unfair clauses, which could infringe upon user rights under European consumer law. This dataset categorizes unfair terms into eight distinct groups: Arbitration, Unilateral Change, Content Removal, Jurisdiction, Choice of Law, Limitation of Liability, Unilateral Termination, and Contract by Using~\cite{lippi2019claudette}. To transform the analysis into a binary classification problem, we re-labelled each sentence as either ``unfair'' if it contains any type of the identified unfair terms, or ``not unfair' if it does not fall into these categories.

\section{Experiments}\label{sec:experiments}
We performed the experiments on a set of binary classification tasks on datasets from various domains, as described in the previous section.

To generate the secondary questions, we employed a large language model. Prompting it only once, we obtained a set of $n$ secondary questions, which we accepted as provided, without any selection or modification. More specifically, we used the following prompt for creating all secondary questions:
\begin{small}
\begin{verbatim}
    Return {n} yes/no questions that would be
    useful to ask if you were trying to 
    determine the answer to the following 
    question: "{primary_question}"
\end{verbatim}
\end{small}
where $n$ is the number of additional questions we want to generate and \verb|primary_question| is the primary question used to obtain the main information from the document. Note that in all our experiments $n=4$. That means that for each document we use one primary and four secondary questions that are treated equally when prompting the LLM. Thus, for each document we collect five answers from the LLM that are then verbalized (assigned a numerical value) in the next step. 
To generate the secondary questions for all our experiments we use OpenAI's \verb|gpt-4-0125-preview| model. To collect the answers in our experiments we use two generations of OpenAI's models: \verb|gpt-4-0125-preview|~\cite{achiam2023gpt} and \verb|gpt-3.5-turbo-0125|~\cite{brown2020language}.

To choose the best classifier, we train several different classification algorithms. These include K-Nearest Neighbors, Decision Trees, Random Forest, Gaussian Naive Bayes, Multinomial Naive Bayes, AdaBoost, and XGBoost. We use a 5-fold cross-validation on our training data and also perform a grid search to fine-tune the parameters for each classifier. After training, we test them on a hold-out test set and choose the classifier that gives us the highest Micro F1 score ($\mu F1$). Note that one can also adjust the training process to optimize for a specific performance metric if needed for a particular application. To perform these experiments, we used the \verb|scikit-learn| library in Python.

Micro F1 score is particularly useful in datasets where some classes are significantly underrepresented, and where traditional metrics might give a misleading picture of model performance. It treats every instance as equally important, thereby giving a more accurate measure of the model’s performance across the board. To calculate the $\mu F1$, we use the following formula:

\begin{equation}
    \mu F1 = \frac{2\mu Precision \times \mu Recall}{\mu Precision + \mu Recall}
    \label{eq:microf1}
\end{equation}

where 

\begin{equation}
    \mu Precision = \frac{\sum{TP_{i}}}{\sum{TP_{i} + \sum{FP_i}}}
\end{equation}

\begin{equation}
    \mu Recall = \frac{\sum{TP_{i}}}{\sum{TP_{i} + \sum{FN_i}}}
\end{equation}

and $TP$, $FP$ and $FP$ represent number of true positives, false positives and false negatives respectively. 

Additionally, we conducted a sensitivity analysis to enhance our understanding of the relationship between the number of features and the improvement of the $\mu F1$. This analysis helps determine the requisite number of secondary questions to attain a desired $\mu F1$. For each dataset, we started by creating $n=9$ secondary questions and using the \verb|gpt-3.5-turbo-0125| model to generate responses for each sample. The outputs from the large language model were then transformed into 10-dimensional feature vectors. Subsequently, we constructed a series of simple Random Forest classifiers, starting with a single feature and incrementally adding more features up to ten. Given the random selection of features for classification, we repeated the experiment 100 times. We computed the $\mu F1$ for each iteration and dataset. The findings are detailed in Section~\ref{sec:results} and illustrated in Figure~\ref{fig:feature_sensitivity}.

\section{Results}\label{sec:results}
The results of the classification experiments are summarized in Table~\ref{tab:results_summary}. We can see that across all datasets, the ICE-T method consistently surpasses the zero-shot approach in performance for a given language models. Specifically, using the GPT-3.5 model, the average $\mu F1$ for the zero-shot approach is 0.683, but it increases to 0.845 with the ICE-T method. A similar trend is observed with the larger GPT-4 model, where the average F1 score improves from 0.7 using the zero-shot approach to 0.892 with the ICE-T technique. This improvement is not constant across the datasets as we can see a significant variations in performance and in improvements across different tasks.

\begin{table}
    \centering
{\small
\begin{tabular}{lcccc}
\toprule
 & \multicolumn{2}{c}{GPT-3.5} & \multicolumn{2}{c}{GPT-4} \\
 & 0-shot &  ICE-T & 0-shot &    ICE-T \\
\midrule
ABDOMINAL         &                    0.791 &  \textbf{0.814} &                  0.802 &  \textbf{0.884} \\
ADVANCED-CAD      &                    0.640 &  \textbf{0.756} &                  0.791 &  \textbf{0.907} \\
ALCOHOL-ABUSE     &                    0.814 &  \textbf{0.965} &                  0.791 &  \textbf{0.965} \\
ASP-FOR-MI        &                    0.849 &   \textbf{0.86} &                  0.860 &  \textbf{0.895} \\
CREATININE        &                    0.349 &  \textbf{0.721} &                  0.488 &  \textbf{0.872} \\
DIETSUPP-2MOS     &                    0.593 &  \textbf{0.767} &                  0.488 &  \textbf{0.814} \\
DRUG-ABUSE        &                    0.942 &  \textbf{0.977} &                  0.826 &  \textbf{0.977} \\
ENGLISH           &                    0.919 &  \textbf{0.988} &                  0.233 &  \textbf{0.965} \\
HBA1C             &                    0.477 &  \textbf{0.837} &                  0.523 &  \textbf{0.942} \\
MAJOR-DIABETES    &                    0.733 &  \textbf{0.814} &                  0.570 &  \textbf{0.884} \\
MAKES-DECISIONS   &                    0.605 &  \textbf{0.965} &                  0.663 &  \textbf{0.965} \\
MI-6MOS           &                    0.651 &  \textbf{0.919} &                  0.849 &  \textbf{0.953} \\
\midrule 
Catalonia indep.  &                    0.528 &  \textbf{0.579} &                  0.562 &  \textbf{0.604} \\
Climate detection &                    0.702 &    \textbf{0.8} &                  0.912 &  \textbf{0.925} \\
ECtHR             &                    0.853 &  \textbf{0.873} &                  0.861 &  \textbf{0.873} \\
Health advice     &                    0.836 &  \textbf{0.841} &                  0.846 &           0.846 \\
UNFAIR-ToS        &                    0.335 &  \textbf{0.887} &                  0.837 &  \textbf{0.889} \\
\midrule 
\textbf{Average}           &                    0.683 &  \textbf{0.845} &                  0.700 &  \textbf{0.892} \\
\bottomrule
\end{tabular}
}
    \caption{\textbf{Comparison of $\mu F1$ scores between zero-shot setting and ICE-T method.} The values in \textbf{bold} represent the $\mu F1$ score of the winning approach for a specific task and a language model. Horizontal line in the middle splits the clinical trial datasets and other datasets. All tasks solve the binary classification problem.}
    \label{tab:results_summary}
\end{table}

The upper portion of Table~\ref{tab:results_summary} showcases the findings from the clinical trial dataset, as detailed in Section~\ref{sec:data}. The dataset's contents remain consistent across all sub-tasks within this clinical trial dataset, though each sub-task involves a distinct classification criterion based on 12 different criteria. In some sub-tasks, substantial improvements were observed over the zero-shot method. For instance, in the task CREATININE (involving serum creatinine levels exceeding the upper normal limit), the zero-shot method achieved $\mu F1$ of 0.349. In contrast, the ICE-T technique utilizing the same large language model significantly improved this score to 0.721. Similarly, for the task ENGLISH (determining if a patient speaks English) using the larger GPT4 model, the greatest increase noted exceeded 0.733 points, with the zero-shot approach at a $\mu F1$ of 0.233 and the ICE-T technique improving it to 0.966. Analysis of tasks outside the clinical trial dataset revealed varied results, dependent on the specific domain. The task assessing ``Catalonia independence'' presented a notable challenge in the zero-shot setup for both models, barely achieving a $\mu F1$ above 0.5, with no significant improvements noted with the ICE-T technique. 

The task related to the European Court of Human Rights (ECtHR) already exhibited high baseline scores in the zero-shot setting, achieving 0.853 with GPT-3.5 and 0.861 with GPT-4. The application of the ICE-T technique yielded minimal improvement, with both models achieving a $\mu F1$ of 0.873. A similar scenario was observed with the Health advice dataset, where enhancements were negligible.

However, the UNFAIR-ToS task demonstrated significant improvement using the ICE-T approach, particularly with the GPT-3.5 model. Here, the $\mu F1$ score saw a dramatic increase from 0.335 to 0.887. 

\begin{figure}
    \centering
    \includegraphics[width=0.98\linewidth]{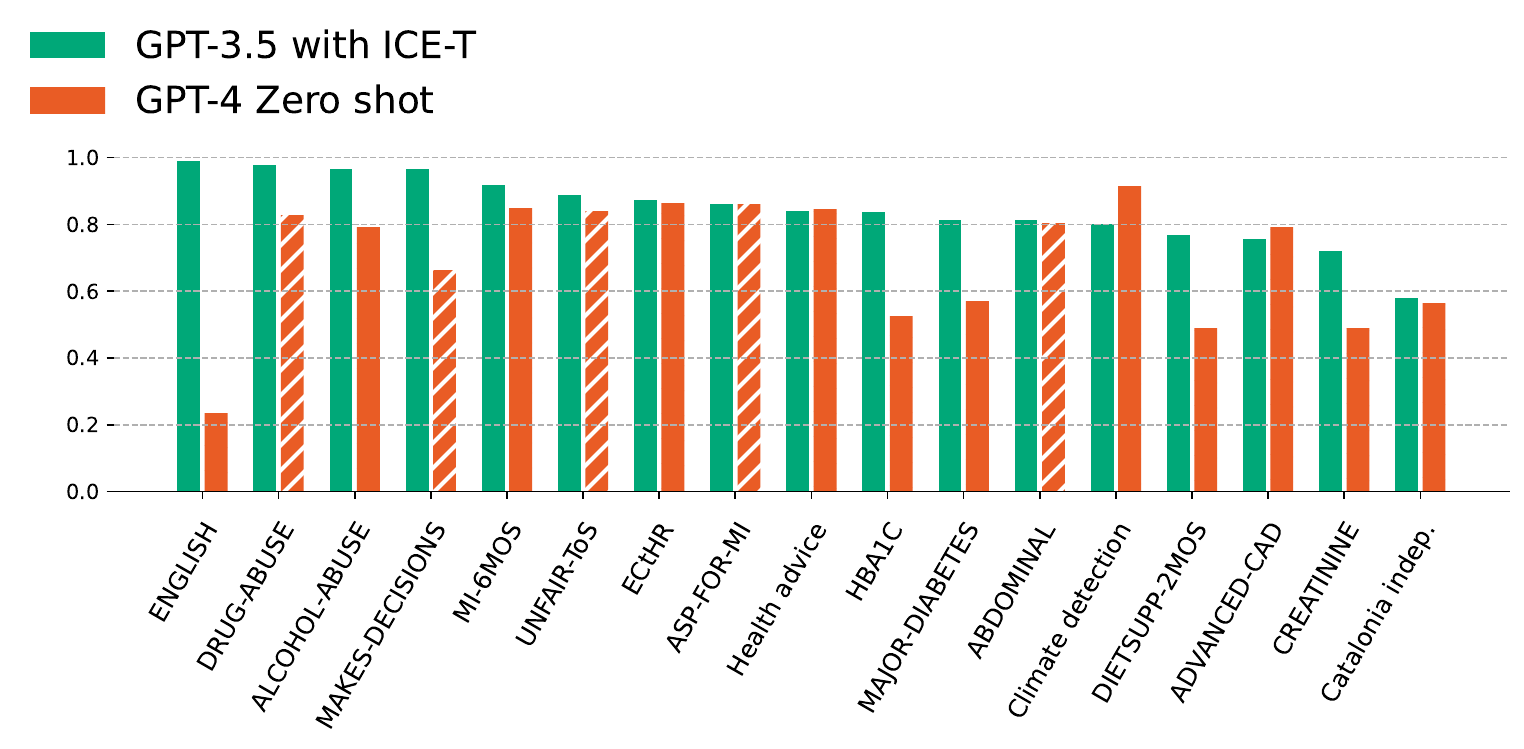}
    \caption{\textbf{Comparative performance of ICE-T-enhanced GPT-3.5 versus zero-shot GPT-4.} The figure illustrates the $\mu F1$ achieved by GPT-3.5 utilizing the ICE-T technique and GPT-4 in a zero-shot setting across multiple datasets.}
    \label{fig:gpt3_vs_gpt4}
\end{figure}

Furthermore, our analysis reveals that the ICE-T technique, when applied to a smaller model, can surpass or match the performance of a larger model that uses the zero-shot approach. In our experiments, we assessed the $\mu F1$ of classification tasks executed by GPT-4 in a zero-shot setting against those performed by GPT-3.5 using the ICE-T technique across various datasets. In nearly all cases, except for two, the ICE-T-enhanced GPT-3.5 either outperformed or equaled the larger GPT-4 model on identical tasks. These findings are depicted in Figure~\ref{fig:gpt3_vs_gpt4}.

We observed a minor variation in performance across different task groups. By categorizing clinical trial tasks into one group and other tasks into another, we observed a comparable average performance improvement when comparing the zero-shot to the ICE-T approach, as detailed in Table~\ref{tab:averages} in Appendix~\ref{sec:additional_results}. This consistency underscores the versatility of the ICE-T method across various domains and tasks.

\begin{figure}
    \centering
    \includegraphics[width=0.98\linewidth]{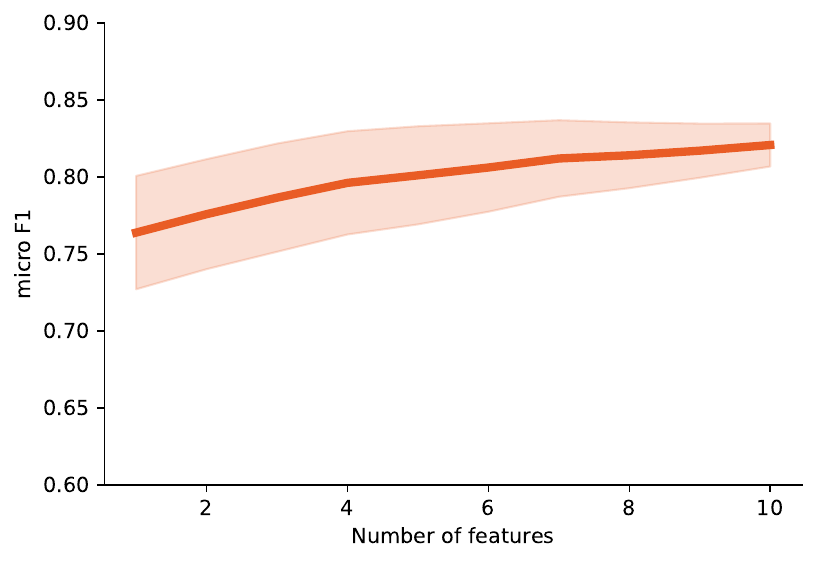}
    \caption{\textbf{Sensitivity Analysis of Feature Count on $\mu F1$ Score.}
The figure illustrates the effect of increasing the number of features (secondary questions) on the $\mu F1$ score across 17 datasets. The solid orange line represents the average $\mu F1$ score, and the shaded area indicates the first standard deviation from the mean across 100 repetitions. The graph demonstrates a consistent improvement in $\mu F1$ as more features are added, with key points of increase highlighted at specific feature counts.}
    \label{fig:feature_sensitivity}
\end{figure}

To explore how the number of features impacts the micro F1 score ($\mu F1$), we conducted an additional sensitivity analysis. The outcomes of this analysis are depicted in Figure~\ref{fig:feature_sensitivity}. This figure illustrates the change of the $\mu F1$ as we incrementally introduce more features (obtained by secondary questions). A solid orange line shows the average $\mu F1$ across all datasets, while the surrounding shaded area indicates one standard deviation from the mean, based on 100 iterations. As anticipated, there is a consistent increase in the micro F1 score with the addition of more secondary questions. On average, adding three secondary questions increases the $\mu F1$ score from 0.76 to 0.80, with further additions raising it to 0.82.

It is important to highlight that this figure averages results from 17 different datasets, using only the Random Forest classifier. Detailed results for each individual task are available in Figure~\ref{fig:feature_sensitivity_per_task} in Appendix~\ref{sec:additional_results}. The use of a single classifier in this analysis was a deliberate choice to isolate the impact of increasing the number of features, thereby minimizing the influence of classifier selection on the results. However, this choice may also limit the generalizability of the findings, as it differs from previous analyses where the optimal classifier was selected for each task.

\section{Discussion}
Our study introduces the Interpretable Cross-Examination Technique (ICE-T), a novel prompting method that integrates LLM responses with traditional classification algorithms to improve the performance on binary classification tasks. This technique addresses key limitations in zero-shot and few-shot learning by employing a structured, multi-prompt approach that transforms qualitative data into quantifiable metrics, thus allowing a small, traditional classifier to effectively make decisions. Our results confirm that ICE-T consistently surpasses zero-shot baselines across multiple datasets and metrics, particularly in scenarios where model interpretability is crucial. This prompting strategy also demonstrates the potential for fully automated, high-performing AI systems accessible even to non-experts.

The ICE-T method has demonstrated its capability to not only enhance performance over the zero-shot approach but also to do so with smaller models that might not perform as well in a zero-shot configuration. For example, the improvement in the CREATININE and ENGLISH tasks within clinical trials data underscores the method's ability to handle domain-specific challenges that require nuanced understanding, which zero-shot configurations typically struggle with. 

\subsection{Implications for Model Interpretability}
A major advantage of the ICE-T approach is its interpretability. By generating a feature vector based on direct responses to structured prompts, experts can trace back the decision-making process, understanding which factors contributed most significantly to the model's classification. This is particularly valuable in fields like medicine and law, where decision rationale is as important as accuracy. The ability to dissect and validate each step of the model's reasoning aligns with the growing demand for transparency in AI applications, ensuring that decisions made by AI systems can be audited and trusted.

Moreover, ICE-T is particularly valuable in situations where fine-tuning models is not viable. Fine-tuned models often suffer from a significant drawback: they lack transparency and become ``black boxes,'' making their decision-making processes obscure. This lack of interpretability is particularly problematic in regulated sectors such as healthcare, law and finance, where it's imperative to comprehend the basis of each decision. ICE-T overcomes these issues by employing a methodology that remains clear and interpretable, avoiding the opaqueness associated with fine-tuned systems.

\subsection{Limitations and Future Work}
Despite its strengths, the ICE-T method has some limitations. The quality of the output heavily relies on the initial set of questions generated for the model to answer. Poorly formulated questions or those that fail to capture the necessary subtleties of the task can limit the effectiveness of this technique. Moreover, the reliance on numerical scoring of textual answers might oversimplify complex answers.This can lead to a loss of nuance, especially when answers are confined to binary outputs.

Future research could explore more sophisticated methods for question generation, perhaps incorporating active learning where the system identifies and prioritizes questions that would most improve its understanding and performance. Additionally, exploring different methods of encoding responses into feature vectors could further enhance the model's accuracy and sensitivity to nuances in text. 

Expanding the scope of ICE-T to tackle problems beyond binary classification could also prove beneficial. Applying this method to multi-class classification tasks or even regression problems could test the adaptability and scalability of the approach, potentially making it more applicable across a wider array of domains. This expansion could lead to significant advancements in the field of machine learning where interpretability and accuracy are crucial.

In conclusion, the ICE-T method presents a promising avenue for enhancing the performance and interpretability of LLMs in binary classification tasks and beyond. By bridging the gap between traditional machine learning techniques and modern LLM capabilities, this approach offers a valuable tool for applications demanding high accuracy and clear reasoning in decision-making processes. Further refinements and adaptations of this technique could significantly impact the deployment of AI in critical sectors, enhancing both the reliability and accountability of automated systems.

\section*{Reproducibility}
The experiment is composed of two primary phases: 1) collecting outputs from OpenAI's ChatGPT models, specifically using either \verb|gpt-4-0125-preview| or \verb|gpt-3.5-turbo-0125|; and 2) verbalizing the answers (converting the responses into numerical form), training classifiers and evaluating their performance on a hold-out test set. 

The code to reproduce the verbalization, classfier training and testing is available on GitHub: https://github.com/gmuric/ICE-T

Due to data usage and confidentiality constraints associated with the clinical trial dataset, we are unable to share the complete working code for the first phase. However, we provide the outputs of the LLMs we obtained. They are available in GitHub repository. We additionally provide pseudo-code that illustrates the extraction of outputs from the language models that can be used to reproduce the first part of the experiment. The complete references to the data used in the experiments are explained in Section~\ref{sec:data}. Additionally, we include a comprehensive list of the questions used to prompt the language models in Appendix~\ref{sec:questions}. The pseudo-code for obtaining the answers from the LLM is presented below:

\begin{algorithm}
\caption{Collecting outputs from LLM}
\begin{algorithmic}
\FOR{each textual document $t$}
    \STATE Prompt LLM with $t$ and corresponding questions \COMMENT{Refer to Section~\ref{sec:prompting_llm}}
    \FOR{each question $q_i$ related to $t$}
        \STATE $a_i \leftarrow$ Output of LLM for $q_i$
    \ENDFOR
    \STATE $A \leftarrow$ Set of all outputs $\{a_1, a_2, \ldots, a_n\}$
    \FOR{each output $a_i$ in $A$}
        \STATE $v_i \leftarrow$ Numerical value of $a_i$ \COMMENT{Refer to Section~\ref{sec:verbalizing}}
    \ENDFOR
    \STATE $V_d \leftarrow$ Feature vector of all $v_i$
\ENDFOR
\end{algorithmic}
\end{algorithm}

Note that due to the stochastic nature of large language models, the outputs may vary with each experiment. While these variations are unlikely to significantly impact the results, minor discrepancies are possible.

\section*{Acknowledgment}
This material is based upon work supported by the Army ASA(ALT) SBIR CCOE under Contract No. W51701-22-C-0035 and the US Air Force under Contract No. FA8750-22-C-0511. Any opinions, findings and conclusions or recommendations expressed in this material are those of the author(s) and do not necessarily reflect the views of the Army ASA(ALT) SBIR CCOE or the US Air Force.
\newpage


%

\balance
\bibliography{acl_latex}

\newpage
\appendix
\section{Questions used in ICE-T method}\label{sec:questions}
This is the list of questions used for each task in ICE-T technique. For each task, there are five questions in total, where the question with $id=0$ is used in a prompt in a zero-shot approach. Other questions are \textit{secondary questions} obtained as explained in Section~\ref{sec:generating_questions}:

\subsubsection*{DRUG-ABUSE}
{\small
\begin{enumerate}
\item[0] Is there any indication in the patient's records of current or past drug abuse?
\item[1] Has the patient been prescribed medication with a high potential for abuse? 
\item[2] Are there any notes in the patient's records indicating substance abuse or dependency issues? 
\item[3] Has the patient previously sought treatment for substance abuse or addiction? 
\item[4] Are there any irregularities in the patient's prescription history that suggest misuse, such as early refill requests?
\end{enumerate}
}

\subsubsection*{ALCOHOL-ABUSE}
{\small
\begin{enumerate}
\item[0] Has the patient consumed alcohol beyond the weekly recommended limits recently?
\item[1] Did the patient consume more than 14 units of alcohol in the past week?
\item[2] Has the patient engaged in binge drinking sessions in the last month?
\item[3] Does the patient frequently consume alcohol on more than 5 days in a week?
\item[4] Has the patient expressed concerns about controlling their alcohol intake recently?
\end{enumerate}
}

\subsubsection*{ENGLISH}
{\small
\begin{enumerate}
\item[0] Does the patient speak English, according to their medical records?
\item[1] Is there a language preference indicated in the patient's medical records?
\item[2] Does the medical record include an interpreter request for non-English languages?
\item[3] Is English listed as the patient's primary language in the medical records?
\item[4] Have previous medical consultations been conducted in English, as noted in the patient's records?
\end{enumerate}
}

\subsubsection*{MAKES-DECISIONS}
{\small
\begin{enumerate}
\item[0] Is there evidence that the patient makes their own medical decisions?
\item[1] Does the patient have a documented history of discussing treatment options with their healthcare provider? 
\item[2] Has the patient previously filled out an advanced directive or a living will?
\item[3] Does the patient regularly attend medical appointments alone? 
\item[4] Have there been instances where the patient explicitly expressed their treatment preferences or declined certain medical interventions?
\end{enumerate}
}

\subsubsection*{ABDOMINAL}
{\small
\begin{enumerate}
\item[0] Is there a record of the patient undergoing intra-abdominal surgery, small or large intestine resection, or experiencing a small bowel obstruction?
\item[1] Has the patient ever undergone any form of intra-abdominal surgery? 
\item[2] Has the patient had a resection of either the small or large intestine? 
\item[3] Is there a record of the patient experiencing a small bowel obstruction? 
\item[4] Has the patient had any surgical intervention related to the digestive system that is not explicitly mentioned above?
\end{enumerate}
}

\subsubsection*{MAJOR-DIABETES}
{\small
\begin{enumerate}
\item[0] Does the patient have any major diabetes-related complications such as amputation, kidney damage, skin conditions, retinopathy, nephropathy, or neuropathy?
\item[1] Has the patient experienced any significant changes in their vision or been diagnosed with retinopathy? 
\item[2] Is there any history of skin conditions, wounds that heal poorly, or any amputations? 
\item[3] Does the patient have a history of kidney damage or been diagnosed with nephropathy? 
\item[4] Has the patient reported any persistent numbness, pain, or tingling in their extremities indicating neuropathy?
\end{enumerate}
}

\subsubsection*{ADVANCED-CAD}
{\small
\begin{enumerate}
\item[0] Is the patient currently taking two or more medications to treat CAD?
\item[1] Is the patient currently being treated for coronary artery disease (CAD)? 
\item[2] Is the patient taking any medication to manage CAD symptoms? 
\item[3] Is the patient prescribed more than one medication specifically for CAD? 
\item[4] Are the medications the patient is taking to treat CAD being taken concurrently?
\end{enumerate}
}

\subsubsection*{MI-6MOS}
{\small
\begin{enumerate}
\item[0] Has the patient had a myocardial infarction within the past 6 months from the most recent record date?
\item[1] Has the patient reported any chest pain or symptoms consistent with a myocardial infarction in the past 6 months? 
\item[2] Has the patient undergone any cardiac diagnostic tests, such as an ECG or troponin levels, in the past 6 months? 
\item[3] Did the patient receive any treatment specifically for myocardial infarction, such as medication, stenting, or bypass surgery, in the past 6 months? 
\item[4] Is there any notation in the patient's medical records of a confirmed diagnosis of myocardial infarction within the past 6 months?
\end{enumerate}
}

\subsubsection*{DIETSUPP-2MOS}
{\small
\begin{enumerate}
\item[0] Has the patient taken any dietary supplements, excluding Vitamin D, in the past 2 months?
\item[1] Have you taken any vitamins or dietary supplements in the past 2 months, besides Vitamin D?
\item[2] Are there any non-prescription supplements you have consumed regularly in the last 60 days?
\item[3] Did you start or continue taking any herbal or nutritional supplements recently, except Vitamin D?
\item[4] Excluding Vitamin D, have you used any health or wellness supplements since two months ago?
\end{enumerate}
}

\subsubsection*{ASP-FOR-MI}
{\small
\begin{enumerate}
\item[0] Is the patient using aspirin to prevent myocardial infarction based on their medical records?
\item[1] Is there a history of cardiovascular disease in the patient's medical records?
\item[2] Has the patient been prescribed aspirin for long-term use?
\item[3] Do the medical records indicate a doctor's recommendation for aspirin to prevent myocardial infarction?
\item[4] Is there any mention of aspirin under the patient's current medications list?
\end{enumerate}
}

\subsubsection*{HBA1C}
{\small
\begin{enumerate}
\item[0] Have any HbA1c test results been listed in the records with a value between 6.5 and 9.5
\item[1] Are there any HbA1c test results listed in the records? 
\item[2] Do any HbA1c test results fall within the 6.5 to 9.5
\item[3] Is the HbA1c value specifically mentioned for each test result? 
\item[4] Have all the HbA1c test results been recorded and updated in the records?
\end{enumerate}
}

\subsubsection*{CREATININE}
{\small
\begin{enumerate}
\item[0] Is there any indication in the patient's records of serum creatinine levels above the upper limit of normal?
\item[1] Has the patient undergone any recent serum creatinine tests? 
\item[2] Are the results of the patient's serum creatinine tests documented in their medical records? 
\item[3] Do the documented serum creatinine levels exceed the established normal range? 
\item[4] Has there been any physician commentary or notes indicating concern over the patient's serum creatinine levels?
\end{enumerate}
}

\subsubsection*{UNFAIR-ToS}
{\small
\begin{enumerate}
\item[0] Does this sentence from company's Terms of Service violate the European Council Directive on unfair terms in consumer contracts?
\item[1] Does the sentence create a significant imbalance between the parties' rights and obligations to the detriment of the consumer?
\item[2] Is the term clearly understandable and transparent to a typical consumer?
\item[3] Does the term determine the main subject matter of the contract, or the appropriateness of the price or remuneration, in a way that is unfair to the consumer?
\item[4] Has the consumer been given an opportunity to negotiate the terms of the contract?
\end{enumerate}
}

\subsubsection*{ECtHR}
{\small
\begin{enumerate}
\item[0] Considering all the facts, did a state breach human rights provisions of the European Convention of Human Rights?
\item[1] Did the state fail to provide adequate protection to an individual or group, thereby violating Article 2 or 3 concerning the right to life and prohibition of torture?
\item[2] Was there any discrimination in the state's actions or laws that contravenes Article 14 or Protocol 12 regarding the prohibition of discrimination?
\item[3] Did the state unlawfully interfere with privacy, family life, freedom of expression, or freedom of assembly and association, as protected under Articles 8 to 11?
\item[4] Was there a failure by the state to ensure a fair trial or access to a court, in violation of Article 6 or 13, concerning the right to a fair trial and the right to an effective remedy?
\end{enumerate}
}

\subsubsection*{Climate detection}
{\small
\begin{enumerate}
\item[0] Is a given paragraph climate-related?
\item[1] Does the paragraph discuss weather, temperature trends, or climate patterns?
\item[2] Does the paragraph mention the impact of human activity on the environment?
\item[3] Does the paragraph reference scientific studies or data on climate change?
\item[4] Does the paragraph address policy or actions taken to mitigate climate impacts?
\end{enumerate}
}

\subsubsection*{Health advice}
{\small
\begin{enumerate}
\item[0] Does the given sentence contain a medical advice?
\item[1] Does the sentence recommend or suggest a specific medication, treatment, or remedy?
\item[2] Does the sentence advise on health behaviors, such as diet, exercise, or sleep patterns?
\item[3] Does the sentence include any wording that implies or directly states a medical diagnosis or prognosis?
\item[4] Does the sentence suggest consulting a healthcare professional or seeking medical attention?
\end{enumerate}
}

\subsubsection*{Catalonia indep.}
{\small
\begin{enumerate}
\item[0] Does the given sentence speak in favor of Catalonia independence?
\item[1] Does the sentence express a positive opinion about Catalonia's political autonomy?
\item[2] Does the sentence criticize the Spanish government's policies towards Catalonia?
\item[3] Does the sentence highlight benefits or advantages of Catalonia being independent?
\item[4] Does the sentence encourage actions or steps towards achieving independence for Catalonia?
\end{enumerate}
}

\section{Additional results}\label{sec:additional_results}
This is the appendix section where additional results are provided.

\begin{table*}[h]
    \centering
{\small
\begin{tabular}{lcccccccc}
\toprule
                  & \multicolumn{2}{c}{$\mu F1$} & \multicolumn{2}{c}{$M F1$} & \multicolumn{2}{c}{$w F1$} & \multicolumn{2}{c}{Sample size} \\
              & 0-shot &       ICE-T & 0-shot &       ICE-T & 0-shot &    ICE-T &       Train & Test \\
\midrule
        ABDOMINAL &              0.791 & \textbf{0.814} &              0.740 & \textbf{0.774} &                 0.775 & \textbf{0.803} &         202 &   86 \\
     ADVANCED-CAD &              0.640 & \textbf{0.756} &              0.593 & \textbf{0.743} &                 0.600 & \textbf{0.746} &         202 &   86 \\
    ALCOHOL-ABUSE &              0.814 & \textbf{0.965} &     \textbf{0.583} &          0.491 &                 0.872 & \textbf{0.948} &         202 &   86 \\
       ASP-FOR-MI &              0.849 &  \textbf{0.86} &      \textbf{0.73} &          0.728 &                 0.834 & \textbf{0.838} &         202 &   86 \\
       CREATININE &              0.349 & \textbf{0.721} &              0.319 & \textbf{0.419} &                 0.256 & \textbf{0.604} &         202 &   86 \\
    DIETSUPP-2MOS &              0.593 & \textbf{0.767} &              0.562 & \textbf{0.765} &                 0.564 & \textbf{0.766} &         202 &   86 \\
       DRUG-ABUSE &              0.942 & \textbf{0.977} &              0.757 & \textbf{0.827} &                 0.954 & \textbf{0.977} &         202 &   86 \\
          ENGLISH &              0.919 & \textbf{0.988} &              0.810 & \textbf{0.978} &                 0.910 & \textbf{0.989} &         202 &   86 \\
            HBA1C &              0.477 & \textbf{0.837} &              0.410 & \textbf{0.834} &                 0.373 & \textbf{0.838} &         202 &   86 \\
   MAJOR-DIABETES &              0.733 & \textbf{0.814} &              0.728 & \textbf{0.814} &                 0.728 & \textbf{0.814} &         202 &   86 \\
  MAKES-DECISIONS &              0.605 & \textbf{0.965} &              0.426 & \textbf{0.491} &                 0.724 & \textbf{0.948} &         202 &   86 \\
          MI-6MOS &              0.651 & \textbf{0.919} &              0.542 & \textbf{0.709} &                 0.724 &  \textbf{0.91} &         202 &   86 \\
 \midrule 
Catalonia indep. &              0.528 & \textbf{0.579} &              0.414 & \textbf{0.536} &                 0.413 & \textbf{0.536} &        2961 & 1145 \\
Climate detection &              0.702 &   \textbf{0.8} &     \textbf{0.673} &          0.444 &        \textbf{0.732} &          0.711 &        1300 &  400 \\
            ECtHR &              0.853 & \textbf{0.873} &      \textbf{0.65} &          0.466 &        \textbf{0.849} &          0.814 &        2384 &  591 \\
    Health advice &              0.836 & \textbf{0.841} &              0.778 & \textbf{0.787} &                 0.830 & \textbf{0.835} &        3470 &  868 \\
       UNFAIR-ToS &              0.335 & \textbf{0.887} &              0.321 &  \textbf{0.47} &                 0.397 & \textbf{0.834} &        3319 &  964 \\
\bottomrule
\multicolumn{9}{l}{{\small $\mu F1$ - Micro F1, $M F1$ - Macro F1, $w F1$ - Weighted F1 }}

\end{tabular}
}
    \caption{Evaluation of ICE-T and Zero-Shot Techniques on GPT-3.5 - This table presents a comparison of F1 micro, macro, and weighted scores for various classification tasks using GPT-3.5. It compares the performance metrics between the zero-shot approach and the ICE-T technique. Additionally, the table includes sample sizes for both training and testing datasets across each task. Notable improvements can be seen in several tasks when utilizing the ICE-T technique over the zero-shot method.}
    \label{tab:gpt3_detailed}
\end{table*}

\begin{table*}[]
    \centering
{\small
\begin{tabular}{lcccccccc}
\toprule
                  & \multicolumn{2}{c}{$\mu F1$} & \multicolumn{2}{c}{$M F1$} & \multicolumn{2}{c}{$w F1$} & \multicolumn{2}{c}{Sample size} \\
              & 0-shot &       ICE-T & 0-shot &       ICE-T & 0-shot &    ICE-T &       Train & Test \\
\midrule
        ABDOMINAL &              0.802 & \textbf{0.884} &              0.757 & \textbf{0.876} &                 0.789 & \textbf{0.885} &         202 &   86 \\
     ADVANCED-CAD &              0.791 & \textbf{0.907} &              0.785 & \textbf{0.906} &                 0.787 & \textbf{0.906} &         202 &   86 \\
    ALCOHOL-ABUSE &              0.791 & \textbf{0.965} &              0.564 & \textbf{0.691} &                 0.856 & \textbf{0.962} &         202 &   86 \\
       ASP-FOR-MI &              0.860 & \textbf{0.895} &              0.770 & \textbf{0.802} &                 0.854 & \textbf{0.881} &         202 &   86 \\
       CREATININE &              0.488 & \textbf{0.872} &              0.487 &  \textbf{0.85} &                 0.477 & \textbf{0.875} &         202 &   86 \\
    DIETSUPP-2MOS &              0.488 & \textbf{0.814} &              0.328 & \textbf{0.814} &                 0.336 & \textbf{0.814} &         202 &   86 \\
       DRUG-ABUSE &              0.826 & \textbf{0.977} &              0.593 & \textbf{0.744} &                 0.879 & \textbf{0.971} &         202 &   86 \\
          ENGLISH &              0.233 & \textbf{0.965} &              0.231 & \textbf{0.925} &                 0.206 & \textbf{0.963} &         202 &   86 \\
            HBA1C &              0.523 & \textbf{0.942} &              0.479 & \textbf{0.941} &                 0.451 & \textbf{0.942} &         202 &   86 \\
   MAJOR-DIABETES &              0.570 & \textbf{0.884} &              0.523 & \textbf{0.884} &                 0.523 & \textbf{0.884} &         202 &   86 \\
  MAKES-DECISIONS &              0.663 & \textbf{0.965} &              0.456 & \textbf{0.691} &                 0.768 & \textbf{0.962} &         202 &   86 \\
          MI-6MOS &              0.849 & \textbf{0.953} &              0.674 & \textbf{0.844} &                 0.868 &  \textbf{0.95} &         202 &   86 \\
 \midrule 
Catalonia indep. &              0.562 & \textbf{0.604} &              0.462 &  \textbf{0.57} &                 0.463 &  \textbf{0.57} &        2961 & 1145 \\
Climate detection &              0.912 & \textbf{0.925} &              0.878 & \textbf{0.896} &                 0.917 & \textbf{0.929} &        1300 &  400 \\
            ECtHR &              0.861 & \textbf{0.873} &     \textbf{0.631} &          0.466 &        \textbf{0.848} &          0.814 &        2384 &  591 \\
    Health advice &              0.846 &          0.846 &              0.766 &          0.766 &                 0.828 &          0.828 &        3470 &  868 \\
       Unfair TOS &              0.837 & \textbf{0.889} &      \textbf{0.63} &          0.489 &        \textbf{0.844} &          0.839 &        3319 &  964 \\
\bottomrule
\multicolumn{9}{l}{{\small $\mu F1$ - Micro F1, $M F1$ - Macro F1, $w F1$ - Weighted F1 }}
\end{tabular}
}
    \caption{Evaluation of ICE-T and Zero-Shot Techniques on GPT-4 - This table presents a comparison of F1 micro, macro, and weighted scores for various classification tasks using GPT-4. It compares the performance metrics between the zero-shot approach and the ICE-T technique. Additionally, the table includes sample sizes for both training and testing datasets across each task. Notable improvements can be seen in several tasks when utilizing the ICE-T technique over the zero-shot method.}
    \label{tab:gpt4_detailed}
\end{table*}

\begin{table*}[h]
    \centering
{\small
\begin{tabular}{llcccc}
\toprule
      &  & \multicolumn{2}{c}{$\mu F1$} & \multicolumn{2}{c}{$MF1$} \\
      &  & 0-shot & ICE-T & 0-shot & ICE-T \\
\midrule
Clinical Trial & GPT3.5 &              0.709 &    0.866 &              0.604 &    0.695 \\
      & GPT4 &              0.673 &    0.915 &              0.560 &    0.803 \\
Other & GPT3.5 &              0.600 &    0.777 &              0.546 &    0.559 \\
      & GPT4 &              0.789 &    0.816 &              0.684 &    0.680 \\
\bottomrule
\end{tabular}
}
    \caption{Comparison of average performance between Zero-Shot and ICE-T methods across Clinical Trial and Other Task Group}
    \label{tab:averages}
\end{table*}

\begin{figure*}[]
    \centering
    \includegraphics[width=0.7\linewidth]{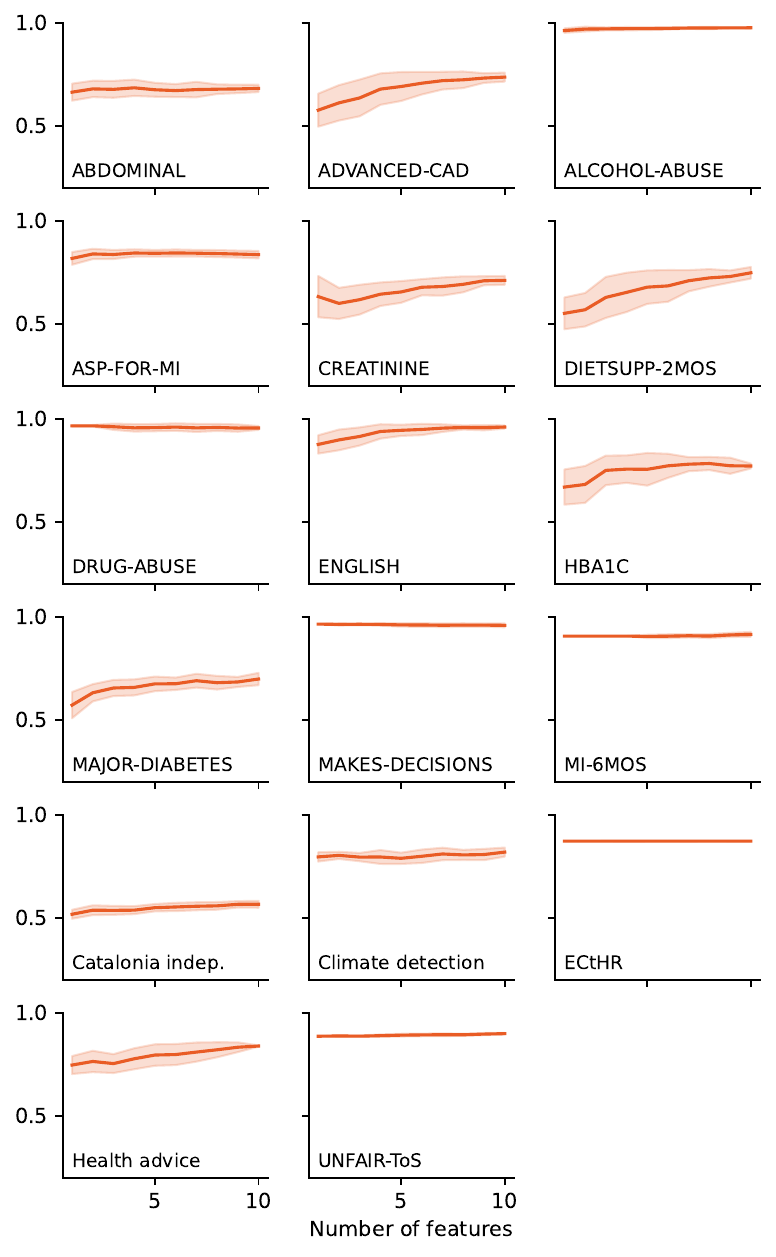}
    \caption{\textbf{Task-Specific Sensitivity Analysis of Feature Count on $\mu F1$ Score.} Detailed view of the changes in the $\mu F1$ score for individual tasks as the number of secondary questions increases. Each plot represents one of the 17 datasets analyzed, showing how the micro F1 score varies with the addition of features. The data underscores the variability in performance improvements across different tasks when using the Random Forest classifier.}
    \label{fig:feature_sensitivity_per_task}
\end{figure*}


\end{document}